\documentclass[letterpaper, 10 pt, conference]{ieeeconf}  

\IEEEoverridecommandlockouts                              

\usepackage{graphicx}
\usepackage{amsmath}
\usepackage{amsfonts}
\usepackage{mathtools}
\usepackage{bbm}
\usepackage{url}
\usepackage{multirow}
\usepackage{booktabs}
\usepackage{dsfont}

\title{\LARGE \bf
Spatial Reasoning from Natural Language Instructions for Robot Manipulation
}

\author{Sagar Gubbi Venkatesh$^{1,2}$, Anirban Biswas$^{3}$, Raviteja Upadrashta$^{1}$,\\ Vikram Srinivasan$^{2}$, Partha Talukdar$^{3}$, and Bharadwaj Amrutur$^{1,2}$
\thanks{$^{1}$Robert Bosch Center for Cyber Physical Systems, Indian Institute of Science, Bangalore 560012, India
        {\tt\small sagar@iisc.ac.in}; {\tt\small ravitejaupadras@iisc.ac.in};  {\tt\small amrutur@iisc.ac.in}}%
\thanks{$^{2}$Department of Electrical and Communication Engineering, Indian Institute of Science, Bangalore 560012, India
        {\tt\small vikram.srinivasan@gmail.com};}%
\thanks{$^{3}$Department of Computational and Data Sciences, Indian Institute of Science, Bangalore 560012, India
        {\tt\small anirbanb@iisc.ac.in}; {\tt\small ppt@iisc.ac.in}}%
\thanks{**This work was supported by Robert Bosch Center for Cyber-Physical Systems, Indian Institute of Science.}
}

\begin{document}

\maketitle
\thispagestyle{empty}
\pagestyle{empty}

\begin{abstract}

Robots that can manipulate objects in unstructured environments and collaborate with humans can benefit immensely by understanding natural language. We propose a pipelined architecture of two stages to perform spatial reasoning on the text input. All the objects in the scene are first localized, and then the instruction for the robot in natural language and the localized co-ordinates are mapped to the \emph{start} and \emph{end} co-ordinates corresponding to the locations where the robot must pick up and place the object respectively. We show that representing the localized objects by quantizing their positions to a binary grid is preferable to representing them as a list of 2D co-ordinates. We also show that attention improves generalization and can overcome biases in the dataset. The proposed method is used to pick-and-place playing cards using a robot arm.

\end{abstract}

\section{INTRODUCTION}

\begin{figure}[!t]
    \centering
    \includegraphics[width=\linewidth]{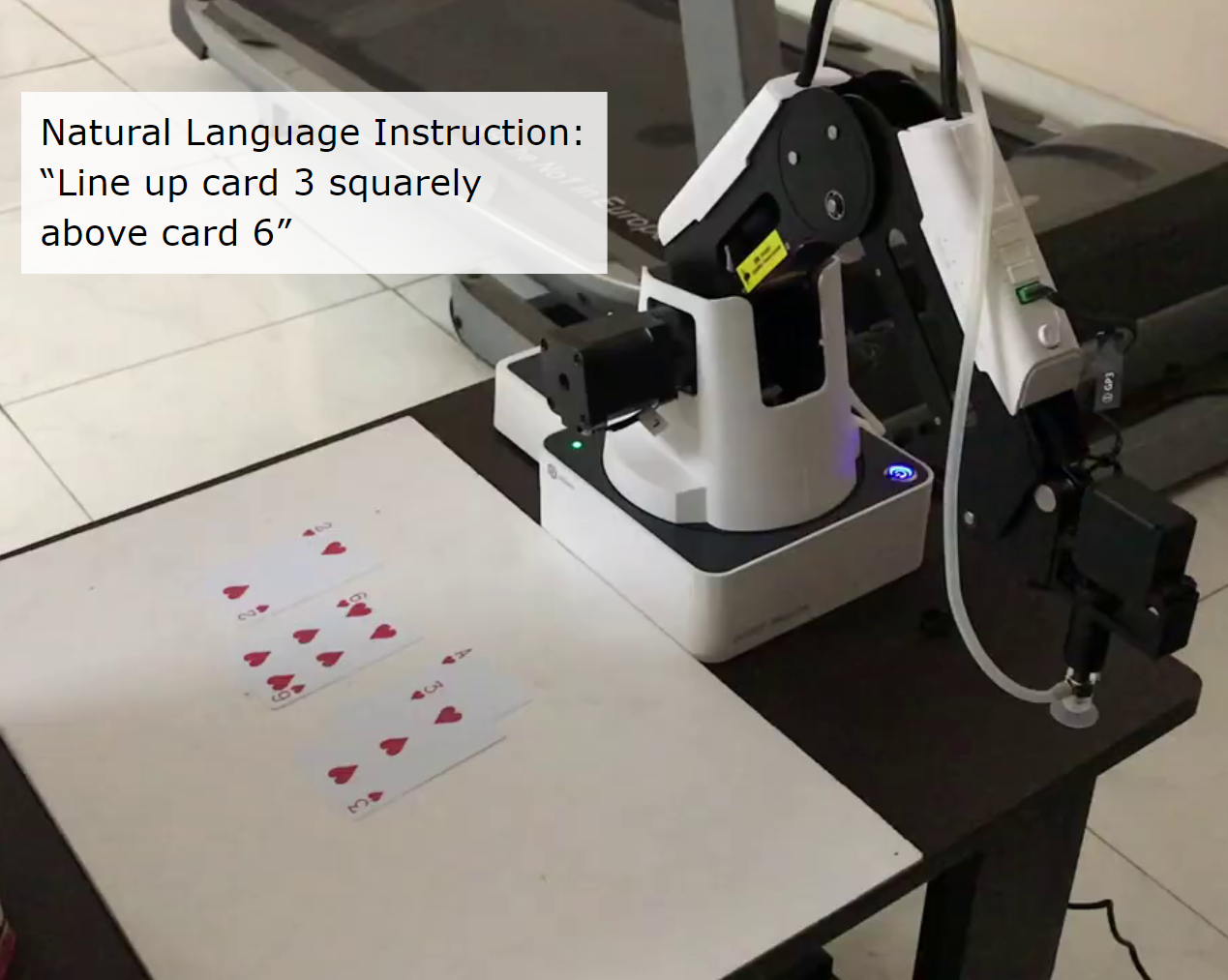}
    \caption{The robot and the cards it must manipulate. The top view of the scene in front of the robot captured from an overhead camera is processed with an object detector to localize all the cards. Based the natural language instruction and the positions of all the cards, the task is to predict the position from which the robot must pick up the card (\emph{start} co-ordinate) and the location at which the card must be placed (\emph{end} co-ordinate).}
    \label{fig:hero}
\end{figure}

This paper addresses the problem of spatial reasoning implicit in natural language instructions to the robot to move objects. Figure~\ref{fig:hero} illustrates a simple model that is representative of this problem. In this example, the objects are the playing cards. The text instruction is of the form ``Line up card 3 squarely above card 6". The robot needs to use this along with the visual input from the camera to infer the \emph{start} co-ordinate from where the robot must pick up the object and the \emph{end} co-ordinate where the object must be placed.

\begin{figure}[!t]
    \centering
    \includegraphics[width=\linewidth]{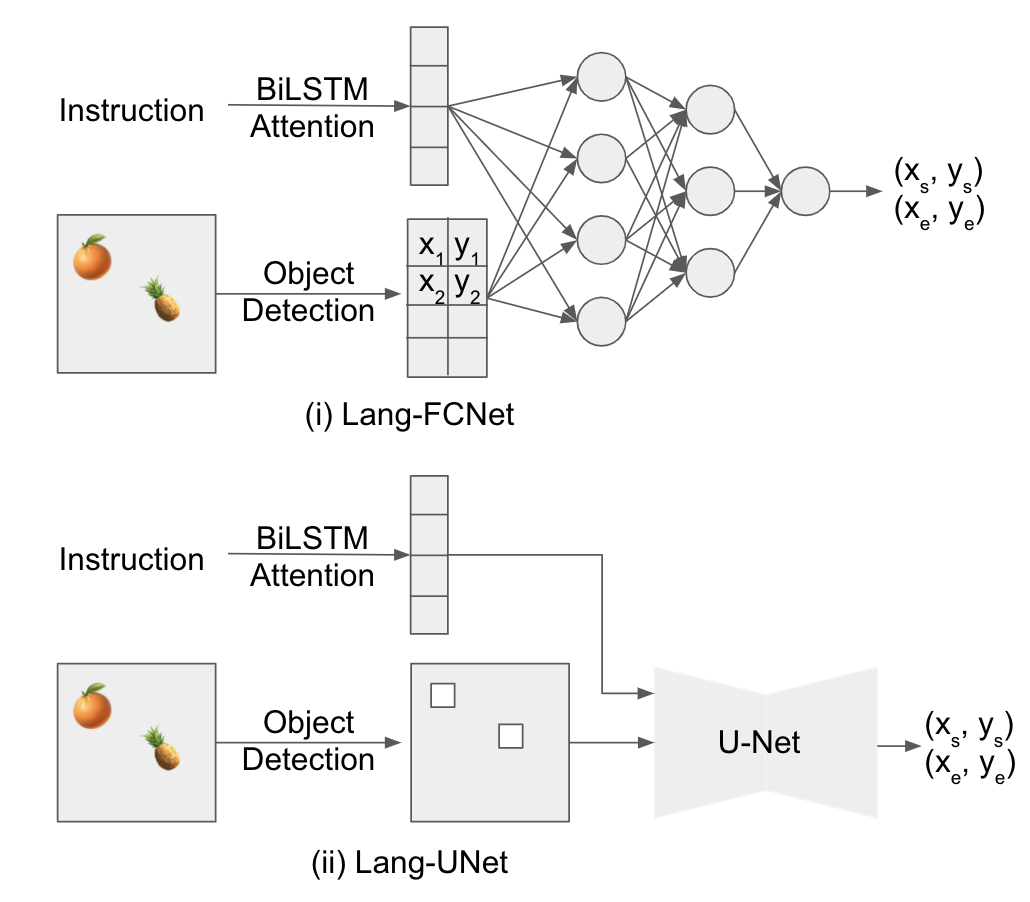}
    \caption{Two architectures for language conditioned localization which use different representations for the output of the object detector.}
    \label{fig:intro_archs}
\end{figure}

One approach is to use an end-to-end network that takes both the image from the camera and the text instruction and directly predicts the physical locations like the \emph{start} and \emph{end} co-ordinates. But such a network must implicitly learn to detect and localize objects, which can be difficult to do from a small dataset. Alternatively, the image from the camera feed can be processed by a separately trained object detector to identify and localize all the objects in the image. The positions of all the objects along with the natural language instruction are then used by a language network to predict the \emph{start} and \emph{end} co-ordinates\cite{bisk-etal-2016-natural}. The approach in \cite{bisk-etal-2016-natural} represents the output of the object detector as a list of 2D co-ordinates indicating the position of the objects in the scene (see Lang-FCNet in Fig.~\ref{fig:intro_archs}(i)). However, this representation has shortcomings which results in poor performance. Hence, we propose an alternative representation for the output of the object detector.

To see why representing the object positions as a list can be sub-optimal, consider the problem of finding ``the second card in a row of cards". If fully connected layers are used to process the coordinate list based output of the object detector\cite{bisk-etal-2016-natural}, the network can overfit to specific locations of the row of cards in the training set, and the network structure is not inherently conducive to generalizing to different positions of the row of cards on the table. To address this, we propose representing the output of the object detector as a binary 2D image with each lit pixel corresponding to an object and using a convolutional network to predict the \emph{start} and \emph{end} positions (see Lang-UNet in Fig.~\ref{fig:intro_archs}(ii)). With this, we expect improved generalization because the convolution operation is, by construction, spatially invariant.

Our contributions:
\begin{itemize}
    \item Object representation in a 2D binary grid via a pre-processing network: We experiment with two different spatial representations and show that instead of representing the localized objects as a list of 2D co-ordinates and processing them with fully connected layers, the spatial reasoning can be improved by representing the detected objects on a 2D binary grid and using a convolutional U-Net to predict the pixels on the grid corresponding to the \emph{start} and \emph{end} positions. 
    \item Multi-head attention and visual grounding: We show that a recurrent network that generates attention for visual grounding generalizes better than a network without attention and can overcome biases in the training dataset.
\end{itemize}

We train the proposed neural network using the blocks dataset\cite{bisk2016towards} and demonstrate the proposed method by deploying it on the Dobot Magician robot arm to pick-and-place playing cards based on text instructions. Our code is open-source\footnote{https://bit.ly/2P3lNGQ}.

\section{RELATED WORK}

    Research on manipulating robots using natural language instructions has gained significant interest. The entire body of related work can be broadly categorized into end-to-end approaches and pipelined approaches. In the end-to-end approaches \cite{misra-etal-2017-mapping}, \cite{mattersim}, \cite{wang2018look}, \cite{fried2018speaker}, \cite{misra-etal-2018-mapping}, \cite{blukis2020few} the robotic agent simultaneously interacts with the surrounding environment while executing natural language instructions and takes a sequence of actions to fulfil its goal. In contrast, the line of work that follows the pipelined approach, \cite{arkin2017contextual}, \cite{paul2016efficient}, \cite{tellex2011understanding}, \cite{arumugam2017accurately} breaks up the task into navigation planning and language grounding processes. Our work is similar to \cite{bisk-etal-2016-natural}, \cite{pivsl2017communication}, that used neural models to localise the scene objects and to ground the spatial relations in unrestricted natural language instructions into a blocks world\cite{bisk2016towards} with complex goal configurations. While their work suffers from poor generalization, we use the attention mechanism \cite{bahdanau2014neural} to solve the problem in a natural way. Recent work in reinforcement learning has demonstrated the usefulness of representing the state information as pixels in a 2D image\cite{laskin2020reinforcement} instead of a list of numbers, which is similar in spirit to this paper although the actual representation is different. 
    
    Earlier works on Human-Robot Interaction have focused on converting a language command with restricted vocabulary and simple actions into a structural form easily understandable by an agent to execute it \cite{klingspor1997human}\cite{mavridis2015review}\cite{dzifcak2009and}. Reinforcement learning based techniques\cite{vogel2010learning}\cite{branavan2009reinforcement} have also been explored for the instruction-following task. The limited action space and little diversity in the language instructions have proven to be non-robust, especially when the instructions are generated by non-experts. Our work addresses these challenges and is able to handle the underlying diversity in the language.
    
    Vision and language grounding are the two important components for an effective human-robot communication through natural language instructions in the context of the surrounding world view. Grounding visual inputs has proven to be extremely essential to many vision tasks like image captioning \cite{hu2016natural}\cite{lu2018neural}, visual question-answering\cite{agrawal2018don}, embodied question-answering\cite{das2018embodied} and vision-language-navigation\cite{misra-etal-2018-mapping}\cite{anderson2019chasing}\cite{mattersim}\cite{ma2019self}\cite{fried2018speaker}. A surfeit of work has been done on grounding natural language instructions \cite{branavan2009reinforcement}\cite{vogel2010learning}\cite{mei2016listen} using variety of techniques like semantic and syntactic parsing\cite{artzi-zettlemoyer-2013-weakly}\cite{macmahon2006walk}, alignment models\cite{andreas-klein-2015-alignment}. Improving language understanding using human-robot dialog\cite{shridhar2018interactive}\cite{thomason2019improving} and commonsense reasoning\cite{chen2019enabling} as well as generation of unambiguous spatial-referring expressions\cite{dougan2019learning} have also been explored. These papers explore different aspects of grounding natural language in vision, whereas our work focuses on spatial reasoning from natural language instructions.
    
    \begin{figure*}[!t]
        \centering
        \includegraphics[width=0.8\linewidth]{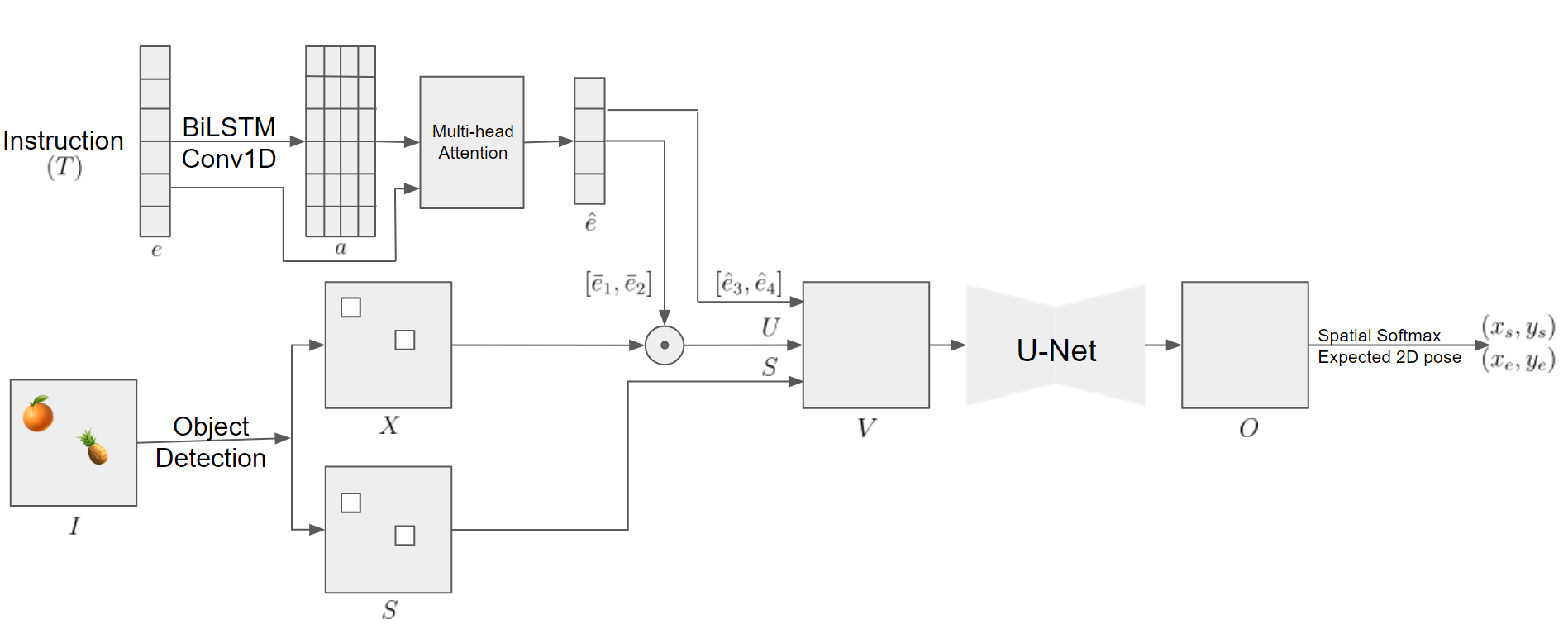}
        \caption{The Lang-UNet model takes the instruction text $T$ and the detected objects $X$ along with their sizes $S$ to predict the \emph{start} and \emph{end} co-ordinates. $\odot$ represents pixel-wise correlation. The U-Net has four Conv5x5(128)-ELU layers followed by four transposed convolutions of the same size and a bottleneck Conv1x1(2) layer. Section~\ref{sec:lang_net} explains the network in more detail.}
        \label{fig:lang-unet}
    \end{figure*}

\section{PROBLEM STATEMENT}
    Given a natural language instruction with embedded spatial cues and an image of the world view, the goal is to understand the instruction in the context of world view and to act in accordance with the spatial cues. For the \textit{pick-and-place} task, the robot must move to the location where the desired object is present, pick it up, and then place it at the goal/target location.  We present a few examples from the datasets in Sec.~\ref{sec:datasets}:
    
    \begin{itemize}
        \item Move block 5 from the top right of box 11 to above box 14 in the middle with a small space.
        \item Place block 5 one and a half columns to the right of block 18.
        \item Pick the first apple from row number one.
        \item Many oranges are placed at random. Pick the biggest orange.
    \end{itemize}
    
    Even though the first two expressions mentioned above differ considerably in their language form, they refer to the same world view and instruct to perform the same action. The model must be robust enough to discern the \emph{start} block (\textit{block 5} in first two examples), choose the correct target anchor (e.g. \textit{block 14} and not block 11, in the first example), recognize the notion of direction (e.g. \textit{right} of block 18 or \textit{above} block 14) and ground the distance information (e.g. \textit{one and a half column} to the right of block 18). The last two representative instructions test the model's ability to understand abstract concepts (e.g. top \textit{row}), reason about object size (e.g. \textit{biggest} orange), ordinality (e.g. \textit{first} apple) and cardinality (e.g. row number \textit{one}).

\section{NETWORK ARCHITECTURE}
\label{sec:lang_net}

    Our proposed \textit{language network} \textbf{Lang-UNet} (Fig.~\ref{fig:lang-unet}) takes as input a natural language instruction and the object positions and sizes from the object detector and finally predicts the \emph{start} $(x_s, y_s)$ and \emph{end} co-ordinates $(x_e, y_e)$. The robot picks the object from \emph{start} location and places it at the \emph{end} location.

    
    The object positions are represented as a binary image $X \in \{0, 1\}^{W\times H \times N_o}$ where $N_o$ is the number of distinct objects. Each object in the scene is represented as a pixel in $X$ by a one-hot vector corresponding to the type of the object. The sizes of the objects are represented in an image $S \in \mathbb{R}^{W\times H}$ with $S_{ij} = 0$ if there is no object at $(i, j)$.
    
    The instruction text $T$ is tokenised with minimal pre-processing (lowercase words, removed punctuation) into a sequence of tokens $\{t_i\}_{i=1}^{N_T}$ and fed into an embedding layer to obtain a vector representation $e_i\in\mathbb{R}^D$ for each token $t_i$. Note that is possible to use BERT\cite{devlin2018bert} to obtain embeddings for the tokens in the instruction, but we chose to learn embeddings from random initialization for easier comparison with a number of previous works, which highlights the benefit of our proposed representation of the object positions and sizes. The token embeddings $\{e_i\}_{i=1}^{N_T}$ are passed through a 2-layer Bi-directional LSTM network. The encoded vector outputs are then passed through a 1-D convolutional network with softmax activation to obtain the attention energies for four attention heads. We denote $a_i^j$ the attention value for $i^{th}$ token and $j^{th}$ attention head. The instruction embeddings are computed as follows: 
\begin{equation}
    \hat{e}_j = \sum_{i=1}^{N_T} a_i^j * e_i
\end{equation}
    
    The first two instruction embeddings are projected to $\mathbb{R}^{N_o}$ using two separate fully connected layers to obtain $\tilde{e}_1$ and $\tilde{e}_2$. Each pixel of $X$ is correlated with $\tilde{e}_1$ to obtain $U^1 \in \mathbb{R}^{W \times H}$ (and likewise $U^2$ is obtained). These two embeddings ``soft-select" appropriate objects in the scene while suppressing the rest.
    
\begin{equation}
    \label{eqn:u-corr}
    U^k_{i, j} = \sum_{l=1}^{N_o} X_{i, j}^l * \tilde{e}^l_k
\end{equation}    
    
    The embeddings $\hat{e}_3$ and $\hat{e}_4$ indicate attributes such as spatial relationships referred to in the instruction. They are repeated $W \times H$ times and appended to $U^1$, $U^2$, and $S$ to get $V = [U^1; U^2; S; \hat{e}_3; \hat{e}_4]$. The image $V$ is passed through the convolutional hourglass network (U-Net) as shown in Fig.~\ref{fig:lang-unet} to obtain $O \in \mathbb{R}^{W \times H \times 2}$. The \emph{start} location $(x_s, y_s)$ and the \emph{end} location $(x_e, y_e)$ are extracted from $O^1$ and $O^2$ respectively by passing it through a spatial-softmax layer.
    
    \begin{equation}
        \hat{O}^k = softmax_{i, j} \big( O^k \big)
        \label{eqn:spatial_softmax}
    \end{equation}
    
    \begin{equation}
        x_s = \sum_{i, j} \hat{O}^1_{i, j} i
       \quad, \quad
        y_s = \sum_{i, j} \hat{O}^1_{i, j} j
        \label{eqn:softargmax}
    \end{equation}
    
    \begin{equation}
        x_e = \sum_{i, j} \hat{O}^2_{i, j} i
       \quad, \quad
        y_e = \sum_{i, j} \hat{O}^2_{i, j} j
        \label{eqn:softargmax2}
    \end{equation}
    
    where $1 \leq i \leq W$ and $1 \leq j \leq H$.
    
    Note that the U-Net structure has no notion of \emph{which} object is at a particular position (notice that it's input size is independent of $N_o$). It is only aware that a particular object ``selected" by the BiLSTM layers via $\tilde{e}_1$ or $\tilde{e}_2$ is present at a location (Eqn.~\ref{eqn:u-corr}). This ensures that the U-Net learns only spatial relationships and not anything specific to an object. So, if the network has learnt to find the position of ``an apple to the left of the banana", it will generalize to ``an orange to the left of the banana".

\section{EXPERIMENTAL RESULTS}

We first evaluate the language network separately on two different datasets. Subsequently, we discuss the performance of the entire pipeline on a real robot arm.

\subsection{Datasets for the Language Network}
\label{sec:datasets}

    To evaluate the language network, we assume that the object positions and sizes are known. We have experimented with two datasets. We use the publicly available Blocks dataset \cite{bisk2016towards}. Additionally we synthesize a diagnostic dataset to test our model performance for more diverse and complicated visual scenarios. We briefly explain the datasets as follows:

    \textbf{Blocks 2D:}
    In the blocks dataset\cite{bisk2016towards}, each sample has a natural language instruction and the positions of all 20 blocks as the input and the labels are the position from which the block must be picked up (\emph{start} position) and the location where the block must be placed (\emph{end} position). A sample instruction: ``Pick up block 9 and place it above block 8". The corpus has a training / development / test distribution of 3712 / 699 / 705 instructions. We allow the \emph{start} and \emph{end} positions to be real valued locations and do not restrict them to be on the grid, but unlike \cite{bisk2018learning} which considers 3D block structures, we retain the 2D world.  We include only those instructions in the blocks dataset that have fewer than 40 tokens in the instruction.
    
    \textbf{Synthetic Dataset:}
    The Blocks dataset has a few limitations: (a) All the blocks are uniquely numbered and only one instance of each block is in the scene, (b) In most of the cases, the instructions are such that the goal location can be obtained by finding the appropriate anchor block and a relative offset direction from a predefined set, and (c) The sizes of the blocks are identical. To diagnose whether the proposed model is capable of reasoning about a variety of other spatial relationships, object attributes (e.g. size), abstract concepts (e.g. row or column) as well as scenes with multiple instances of each object, we build a synthetic dataset. We follow a similar approach proposed in \cite{johnson2017clevr} and generate $\sim$42,000 unique instructions with varied scenes containing objects of sizes randomly chosen between 1.0 to 3.0 and divide it into train / dev / test distribution of 29465 / 4216 / 8416. Each scene contains a maximum of 12 distinct objects and up to a total of 24 objects. Some of the representative templates used to generate instructions are as follows: (i) \textit{Pick the largest / smallest $\langle$obj$\rangle$}, (ii) \textit{Pick the leftmost / rightmost $\langle$obj$\rangle$ from the row of $\langle$obj$\rangle$s}, (iii) \textit{Pick the $\langle$obj\_pos$\rangle$ $\langle$obj$\rangle$ from top row}, (iv) \textit{Pick the $\langle$obj1$\rangle$ above / below / to the left of / to the right of $\langle$obj2$\rangle$}. For this dataset, we predict only the \emph{start} location and ignore the \emph{end} location.

\begin{figure}[!t]
    \centering
    \includegraphics[width=1.0\linewidth]{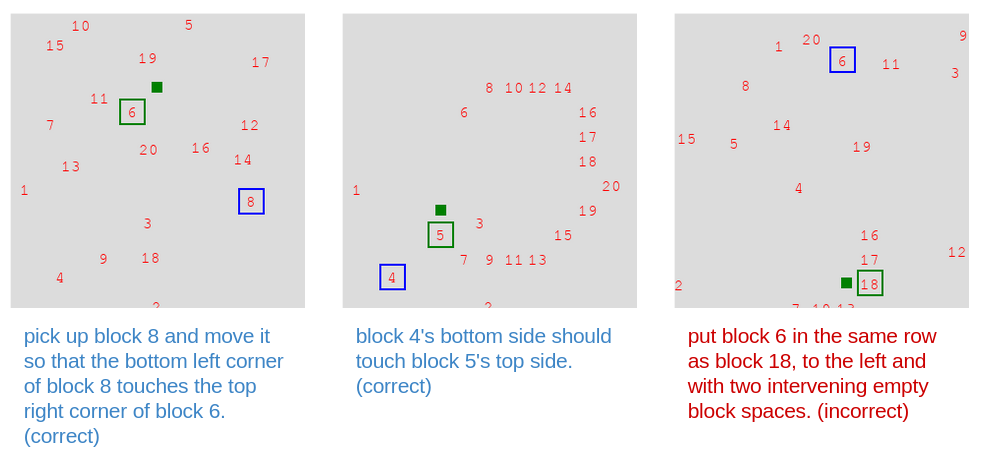}
    \caption{Sample predictions of location of \emph{start}, anchor block (with blue and green squares), and \emph{end} location (green spot) from the \textit{Lang-UNet} for the Blocks dataset.}
    \label{fig:sample_vis}
\end{figure}

\begin{figure}[!t]
    \centering
    \includegraphics[width=0.9\linewidth]{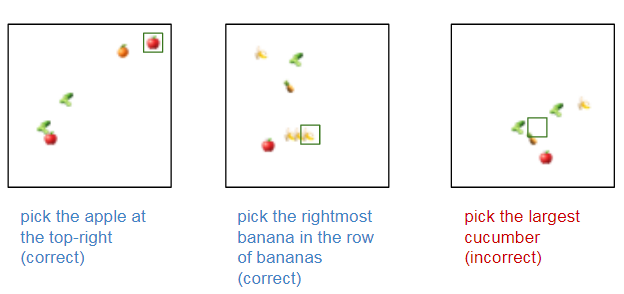}
    \caption{Sample predictions of \textit{Lang-UNet} for the synthetic dataset. The network predicts only the $(x, y)$ location, but to aid visualization, the prediction is represented as a green square box. In the last example, the difference in the sizes of the cucumbers is small (the sizes are randomly chosen when synthesizing the dataset), which causes the network to incorrectly predict the location.}
    \label{fig:sample_vis_synth}
\end{figure}

\subsection{Evaluation Metrics for the Language Network}
We define two evaluation metrics to compare the baseline results quantitatively with ours.

\textbf{Mean Squared Error (MSE):} It is the average over the squared distances between the gold and predicted locations. Define the prediction and gold locations for $i^{th}$ instruction as $L_P$ and $L_G$ respectively. Then $MSE = \frac{\sum_{i=1}^{N}{\left\| L_P^i - L_G^i \right \|}^2}{N}$. The center of the simulated world is at $(0,0)$ and restricted in the range of $[-1,1]$ in both $x$ and $y$ directions.


\textbf{Tolerable Accuracy (TA):} In majority of the real world applications, it is acceptable even if the predicted and the target locations do not exactly match but the distance between them is within a certain tolerable (application-specific) range. To account for this fact, we propose a new metric \textit{Tolerable Accuracy}. A prediction is considered to be correct if the distance error (in both $x$ and $y$ directions in simulated world) is less than a tolerance value \textit{tol}. We count the number of correct prediction instances out of the total $N$ instructions to evaluate TA. Mathematically,  
$TA = \frac{\sum_{i=1}^{N}{\mathds{1}\{|L_P^i - L_G^i|_{x,y} < tol \}}}{N}$. $\mathds{1}\{z\}$ is an indicator function and it's value is $1$ when $z$ is \textit{true}, otherwise $0$.


\subsection{Baseline Algorithms for the Language Network}\label{sec:baseline_algos}

    We compare our proposed model with the following baseline algorithms:

    \textbf{Center:} This model assumes complete knowledge about the \emph{start} location and places the block at the middle of the table.   

    \textbf{Random:} The Random baseline decides both the \emph{start} and \emph{end} locations to be random. The \textit{Center} and \textit{Random} are two simple baselines taken from \cite{bisk-etal-2016-natural}. 

    \textbf{LSTM \cite{johnson2017clevr}:} The word embeddings of the instructions obtained from an embedding layer is passed through a word-level multi-stage LSTM model. The output from the last layer of LSTM is the instruction embedding and passed through a MLP to obtain the final outputs. This model does not use the image information, hence can identify object positions only through the presence of any bias in the instructions. We included this baseline to show that the dataset is not biased in such a way that the instruction alone can predict the \emph{start} and \emph{end} positions.

    \textbf{LSTM+CNN \cite{johnson2017clevr}:} This is an end-to-end approach that takes the image and the text instruction as input and directly predicts the \emph{start} and \emph{end} positions. As above, the instruction embedding is obtained from the last-layer hidden state of LSTM and the image is encoded using CNN features. They are concatenated and fed into a MLP to get the prediction of the \emph{start} and \emph{end} locations.

    \textbf{LSTM+CNN+SA \cite{johnson2017clevr}:} The encodings for image and instruction are generated as above and then soft spatial attention is employed to get the final representation. A MLP takes this as input and produces the required output.

    \textbf{LSTM+UNet \cite{misra-etal-2018-mapping}:}
    The Ling-UNet\cite{misra-etal-2018-mapping} is an end-to-end architecture that takes the RGB pixels from the camera and the instruction text and directly regresses the \emph{start} and \emph{end} positions. This is similar to the \textbf{LSTM+CNN} architecture above, but a UNet is used instead of the CNN and vectors derived from the instruction text using the LSTM are convolved with skip connection activations of the UNet. Similar to the proposed architecture, the \emph{start} and \emph{end} positions are extracted from the last layer of the UNet using the spatial-softmax layer.

    \textbf{RNN-NoAttn-NoGround \cite{bisk-etal-2016-natural}:} The model architecture has a single-layer RNN at its heart. It takes as input the instruction and predicts the \emph{start} object, anchor object, and chooses from 8 pre-defined offsets corresponding to the 8 adjacent positions (right, bottom-right, etc.). The \emph{start} position prediction is simply the position of the predicted object to be picked up. The \emph{end} position is obtained by adding the position of the predicted anchor object and the predicted offset. Note that this model is not grounded because the predictions of the \emph{start}, anchor, and offset are invariant to the positions of the objects in the scene and depend only on the natural language instruction. Furthermore, it cannot distinguish between two or more instances of the same object in the scene. For fair comparison, we train this model on our dataset.

    \textbf{LangNet-Attn-NoGround:} We extend the model in Bisk et. al \cite{bisk-etal-2016-natural} by introducing an attention mechanism over a multi-layer BiLSTM.
    
    \textbf{Lang-FCNet:} This model differs from the proposed Lang-UNet model in that object positions are represented as a list of 2D points rather than as a  binary image, and they are passed through fully connected layers rather than the convolutional hourglass network (U-Net).
    
    

\subsection{Results for Language Network}\label{sec:results}

\begin{figure}[!t]
    \centering
    \includegraphics[width=0.75\linewidth]{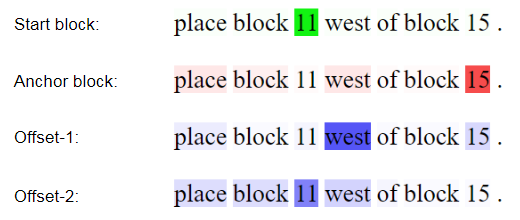}
    \caption{Visualization of attention in \textit{Lang-UNet}. The darker colors indicate higher attention weights for that token. Two attention heads indicate the \textit{Offset} ($\hat{e}_3$ and $\hat{e}_4$) and the other two for the objects corresponding to the \emph{start} and \emph{end} predictions ($\hat{e}_1$ and $\hat{e}_2$).}
    \label{fig:sample_attn}
\end{figure}

\begin{table*}[!t]

\begin{center}
{
\begin{tabular}{clcccccc}
\toprule

& \multirow{3}{*}{Model} & \multicolumn{4}{c}{Blocks 2D} & \multicolumn{2}{c}{Synthetic} \\

\cmidrule{3-6} \cmidrule{7-8}

& & \multicolumn{2}{c}{Start} & \multicolumn{2}{c}{End} & \multicolumn{2}{c}{Start} \\
& & MSE & TA (\%) & MSE & TA (\%) & MSE & TA (\%) \\[1mm]
 
\midrule

\multirow{7}{*}{\rotatebox[origin=c]{90}{Baselines}} & Center & 0.3130 & 1.28 & 0.3130 & 1.28 & 0.6018 & 0.02 \\
& Random & 1.4379 & 0.28 & 1.0118 & 0.14 & 1.2539 & 0.25 \\


\cmidrule{2-8}

& LSTM & 1.1752  & 0.28  & 0.3026 & 1.70 & 0.1694 & 2.82 \\

& LSTM+CNN & 0.8501  & 0.00  & 0.3154 & 0.14 & 0.0302 & 28.92 \\

& LSTM+CNN+SA & 1.0789 & 0.00 & 0.3706 & 4.40 & 0.0125 & 44.89 \\

\cmidrule{2-8}

&  LSTM+UNet \cite{misra-etal-2018-mapping} & 0.1502 & 26.24 & 0.1436 & 23.55 & 0.0496 & 14.50 \\

&  RNN-NoAttn-NoGround \cite{bisk-etal-2016-natural} & \textbf{0.0067} & 96.88 & 0.0392 & 54.33 & 0.0619 & 2.89 \\

\midrule

\multirow{3}{*}{\rotatebox[origin=c]{90}{Ours}} 

& LSTM-Attn-NoGround & 0.0084 & \textbf{98.51} & \textbf{0.0306} & \textbf{66.85} & 0.0849 & 9.49 \\
& Lang-FCNet & 0.0109 & 97.73 & 0.0713 & 61.56 & 0.0145 & 59.88 \\

\cmidrule{2-8}

& Lang-UNet & 0.0097 & 97.00 & 0.0491 & 60.07 & \textbf{0.0031} & \textbf{95.99} \\

\bottomrule

\end{tabular}
}

\caption{Comparison of the performance of various baselines and our models. In all the experiments and models, \textit{tol} is set to 0.05 to measure TA. The \textit{Baselines} are explained in Sec.~\ref{sec:baseline_algos} and \textit{Lang-UNet} in Sec.~\ref{sec:lang_net}. We discuss more about the results in Section~\ref{sec:results}.}
\label{tab:results}
\end{center}

\end{table*}

\textbf{Better generalisation with Attention:} The language network is trained on the blocks dataset using the Adam optimizer with mean absolute error loss, with learning rate 1e-3, and weight decay 1e-9. A few sample predictions of the \textit{Lang-UNet} (BiLSTM model with attention) model are visualized in Figs.~\ref{fig:sample_vis} and \ref{fig:sample_vis_synth}. For one example, the attention weights for the different tokens of the instruction when predicting the \emph{start}, anchor, and offsets are shown in Fig.~\ref{fig:sample_attn}. Table~\ref{tab:results} compares the performance of the proposed approach with the baselines. We observe that the attention-based models - \textit{LSTM-Attn-NoGround} and \textit{Lang-UNet} perform significantly better than the \textit{RNN-NoAttn-NoGround} model that has no attention component, especially for \emph{end} co-ordinate prediction. Fig.~\ref{fig:sample_attn} shows the attention mechanism is able to attend on the correct offset and target block and intuitively explains the reason behind improved performance for \emph{end} co-ordinate prediction. Note that the accuracy of predicting the \emph{end} location is worse than for the \emph{start} location. This is because in most textual instructions, the \emph{start} location is simple and unambiguous (``place block 4..." or ``pick up block 3..."), whereas the target is more complex (last example in Fig.~\ref{fig:sample_vis}) and sometimes ambiguous (``the 14th block moved next to the 12th block"). It also suggests why attention is more important for predicting the \emph{end} location than the \emph{start} in case of Blocks dataset.

\textbf{Mitigating the effect of bias through Attention:} We noticed that the Blocks dataset is biased with some block numbers more frequently being associated with some offsets (such as ``north of") than others. Because of this, the \textit{RNN-NoAttn-NoGround} model overfits and always predicts the same offset when some block numbers are present in the instruction and ignore the actual content of the text. In contrast, the \textit{LSTM-Attn-NoGround} model and the \textit{Lang-UNet} are forced to attend to the offset token in the instruction and gets it right. For example, all the models correctly predict the output for ``Move block 4 above block 5". But, when the block number is changed to ``Move block 11 above block 5", only  \textit{Lang-UNet} and \textit{LSTM-Attn-NoGround} make the correct prediction. To quantify this, we selected 20 simple examples such as the above example from the validation set. The \textit{Lang-UNet}, \textit{RNN-NoAttn-NoGround} models correctly predicted 19 and 19 examples respectively. But when we randomized the block numbers in those instructions, the number of correct predictions were 19 and 6 respectively. Attention was necessary to retain performance and demonstrates its usefulness in such biased datasets.

\textbf{Visual grounding is essential for diverse and complex data:} We note that the ungrounded models which use the natural language instruction alone and do not use object positions perform reasonably on the Blocks dataset. However, they perform poorly on the synthetic dataset because it is not possible to predict the correct position for an instruction such as ``Pick the apple to the left of the orange" without actually using the object positions. The proposed \textit{Lang-UNet} peforms well on both the Blocks and the synthetic datset, but slightly underperforms \textit{RNN-NoAttn-NoGround} on the Blocks dataset due to the quantization in representing the object positions. Moreover, the \textit{LSTM-Attn-NoGround} and \textit{RNN-NoAttn-NoGround} models use hard-coded offsets ($\{0, 0.166, -0.166\}$ in both X and Y directions) that are added to the position of the anchor object to predict the \emph{end} position and are thus specialized to the Blocks dataset, whereas the proposed \textit{LangUNet} does not use such hard-coded offsets.

\textbf{Benefits of Grid representation over List:} The \textit{Lang-FCNet} takes the output from the localisation network as a list of 2D points, whereas the \textit{Lang-UNet} considers a 2D binary grid representation as explained in Sec.~\ref{sec:lang_net}. Fig.~\ref{fig:intro_archs} depicts the differences between the output formats from the localisation network. On examples such as ``Pick up the banana from the row of bananas", \textit{Lang-UNet} is successful in predicting the location of the banana because the convolutional layers in the U-Net help in recognizing ``rows", whereas \textit{Lang-FCNet} performs poorly.  From Table~\ref{tab:results}, we infer that the performance improvement in \textit{Lang-UNet} over \textit{Lang-FCNet}, particularly for the synthetic dataset, is due to the binary grid representation because it provides the model a better way to understand the object positions and the relative spatial relationships amongst each other compared to a list of 2D coordinates. Our empirical evaluation in Table~\ref{tab:results} also suggests the superiority of the pipelined approaches (\textit{Lang-FCNet} and \textit{Lang-UNet}) over the end-to-end models (\textit{LSTM+CNN} and \textit{LSTM+CNN+SA}).

\subsection{Demonstration on the Robot Arm}

We demonstrate the complete pipeline using a Dobot Magician robot arm (Fig.~\ref{fig:hero}). Playing cards are placed at random positions in front of the robot on the table. A camera mounted overhead captures the the top view of the table. The position of the cards is obtained using an object detector\cite{liu2016ssd} that is fine tuned to detect playing cards. Based on the instruction and the positions of the playing cards on the table, the robot picks-and-places a card. Note that while we can switch out the playing cards and instead detect other objects, it is necessary to know a priori what objects will be referenced in the instruction text and to fine-tune the object detector to detect those objects. Out of 15 trials with playing cards, the robot successfully picks the right card in all the trials. In 14 cases, the card is placed within 1~cm of the target. In one case, the localization of the anchor is off by more than 1~cm. A video of the robot in operation is available at \url{https://youtu.be/bfmDC-zoCFc}.

\section{CONCLUSIONS}
In this paper, we have illustrated the advantages of a pipelined approach to manipulating objects based on natural language instructions. We propose having a separately trained object detector followed by a language network that is responsible for predicting the \emph{start} and \emph{end} positions to pick-and-place objects based on the natural language instruction. We show that representing the positions of the detected objects on a 2D binary grid and processing them with a convolutional hourglass network results in much better performance than representing them as a list of 2D co-ordinates and processing with fully connected layers. We also show that attention improves the generalization, especially when the training data is biased. 

\addtolength{\textheight}{-1cm}   




\section*{ACKNOWLEDGMENT}
We would like to thank the Robert Bosch Center for CyberPhysical Systems for funding support.


\bibliographystyle{IEEEtran}
\bibliography{IEEEabrv,mybibfile}

\end{document}